\documentclass[final]{l4dc2026}  
\usepackage{hyperref}       
\usepackage{url}            
\usepackage{booktabs}       
\usepackage{amsfonts}       
\usepackage{nicefrac}       
\usepackage{microtype}      
\usepackage{xcolor}         
\usepackage{amsmath}
\usepackage{graphicx}
\usepackage{algorithm}
\usepackage[noend]{algpseudocode}      
\algrenewcommand\algorithmiccomment[1]{\hfill// #1}

\usepackage[font=small,labelfont=bf,skip = 2pt]{caption}
\usepackage{wrapfig}
\usepackage{multirow}

\usepackage{array} 

\title[Constraint-Aware Adaptive Action Scaling]{Constraint-Aware Reinforcement Learning via Adaptive Action Scaling}
\usepackage{times}

\author{%
 \Name{Murad Dawood} \Email{dawood@cs.uni-bonn.de}\\
 \AND 
 \Name{Usama Ahmed Siddiquie} \Email{s46usidd@uni-bonn.de}\\
  \AND
  \Name{Shahram Khorshidi} \Email{khorshidi@cs.uni-bonn.de}\\
  \AND
 \Name{Maren Bennewitz} \Email{maren@cs.uni-bonn.de}\\
 \addr Humanoid Robots Lab, University of Bonn, Germany%
}
\begin{document}

\maketitle

\begin{abstract}
Safe reinforcement learning (RL) seeks to mitigate unsafe behaviors that arise from exploration during training by reducing constraint violations while maintaining task performance. Existing approaches typically rely on a single policy to jointly optimize reward and safety, which can cause instability due to conflicting objectives, or they use external safety filters that override actions and require prior system knowledge.
In this paper, we propose a modular cost-aware regulator that scales the agent’s actions based on predicted constraint violations, preserving exploration through smooth action modulation rather than overriding the policy. The regulator is trained to minimize constraint violations while avoiding degenerate suppression of actions.
Our approach integrates seamlessly with off-policy RL methods such as SAC and TD3, and achieves state-of-the-art return-to-cost ratios on Safety Gym locomotion tasks with sparse costs, reducing constraint violations by up to 126 times while increasing returns by over an order of magnitude compared to prior methods.

\end{abstract}
\section{Introduction}
\begin{wrapfigure}{r}{0.5\linewidth}
  \centering
  \vspace{-19pt} 
  \includegraphics[width=\linewidth]{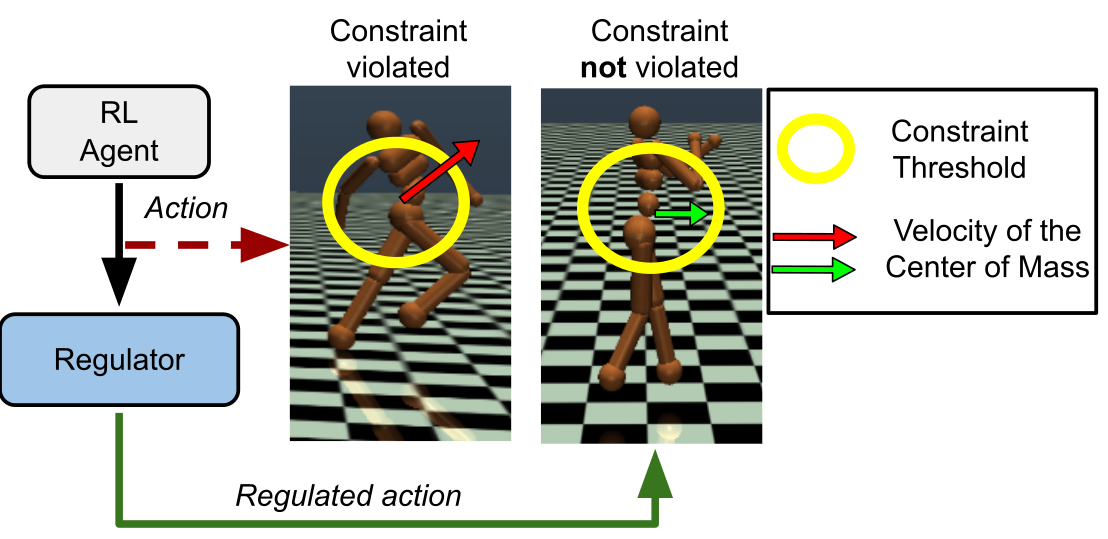}
  \caption{\textbf{Overview of cost-aware action scaling.} 
  The RL agent proposes an action that would result in the center of mass (COM) exceeding the velocity threshold (left).
  The regulator (blue) scales the action, keeping the velocity of the COM within the safe zone while allowing progress on the task. 
  The yellow circles highlight the velocity threshold for the COM.}
  \label{fig:cover}
  \vspace{-8pt} 
\end{wrapfigure}
Reinforcement Learning (RL) has demonstrated remarkable success across a range of domains, including Atari games~\cite{mnih2015human}, robotics~\cite{gu2017deep, wu2023daydreamer}, and long-horizon strategy games \cite{silver2017mastering, vinyals2019grandmaster}. This success is significantly facilitated by exploratory behavior, which allows agents to discover effective behaviors. However, such exploratory behaviors often lead to the violation of system constraints. While such violations are tolerable in simulation and games with free resets, they pose serious risks in real-world applications \cite{amodei2016concrete}. Violating safety constraints can lead to irreversible damage or system failure. To address this issue, Safe Reinforcement Learning (Safe RL) \cite{garcia2015comprehensive} aims to minimize constraint violations during both training and deployment.

Safe RL methods can be broadly categorized into two groups: \emph{safe exploration} and \emph{constrained~RL}. Safe exploration techniques aim to prevent the agent from taking actions that violate safety constraints. These methods typically rely on prior knowledge of the system dynamics and feasible safe states to construct control barrier functions~\cite{ames2019control, dai2023safe, zhang2023spatial}, or model predictive shields~\cite{banerjee2024dynamic, dawood2025l4dc, agha2024exploring}. Although effective, their applicability is limited by the need for detailed prior information about the system dynamics, an assumption that often does not hold in early learning stages or tasks where system dynamics are unknown.

Constrained RL allows the agent to learn both reward and cost signals online, without requiring knowledge of the system dynamics. The agent is trained to maximize cumulative rewards while minimizing constraint violations. Common approaches include Lagrangian-based methods~\cite{achiam2017constrained,tessler2018reward, ray2019benchmarking, stooke2020responsive}, and budget-based methods~\cite{sootla2022saute,sootla2022enhancing}.
However, a core limitation of these methods is the difficulty of balancing reward and cost within a single policy. Conflicting gradients can cause the agent to behave either too conservatively or unsafely, leading to instability, constraint violations, or poor performance~\cite{stooke2020responsive, navon2022multi}.

In contrast to prior work, we propose a modular alternative: instead of overriding actions or jointly optimizing conflicting objectives, we scale actions based on the expected cost of future constraint violations while preserving the policy’s task-directed behavior (see Fig.~\ref{fig:cover}).The architecture consists of a reward-maximizing task agent and a regulator network guided by twin cost critics to conservatively estimate constraint violations. The regulator applies element-wise scaling to attenuate risky actions, enforcing safety without requiring prior knowledge of dynamics or compromising exploration.
Although our approach resembles safe exploration in formulation, we do not require prior knowledge of the system dynamics, and we do not override the task agent's actions, thereby preserving both exploration and safety without external overrides.

We evaluate our approach on several dynamical systems from Safety Gymnasium~\cite{ji2023safety}. Our method achieves the highest Return-to-Cost (RC) ratio \cite{thananjeyan2021recovery}, reducing constraint violations by up to 126~times over recent safe RL baselines \cite{stooke2020responsive, sootla2022enhancing, yu2022towards, ganai2023iterative, kim2024spectral}.
In summary, our contributions are: (i) We propose a modular safe RL framework that decouples reward maximization and safety enforcement via a cost-aware regulator that scales actions based on predicted violations. (ii) Our model-free approach integrates seamlessly into standard off-policy RL pipelines such as SAC~\cite{haarnoja2018soft} and TD3~\cite{fujimoto2018addressing}, improving safety without compromising exploration. (iii) We achieve state-of-the-art performance on safety benchmarks, with up to 126~times fewer constraint violations and the highest RC ratios across tasks.
\vspace{-7pt}

\section{Related Work}

Existing approaches fall into two categories: \emph{safe exploration methods}, which prevent unsafe actions, and \emph{constrained RL methods}, which embed cost objectives directly into policy optimization.

\textbf{Safe Exploration Methods.}
Early methods, \cite{sui2015safe}, employed uncertainty modeling through Gaussian Processes to restrict exploration and ensure safety. Later approaches introduced safety layers~\cite{dalal2018safe, sheebaelhamd2021safe} and predictive safety filters~\cite{banerjee2024dynamic, dawood2025ral,agha2024exploring,tian2024reinforcement} that prevent risky actions based on pre-trained layers or model predictive control (MPC). Control Barrier Function~(CBF)-based strategies~\cite{ames2019control,dai2023safe,zhang2023spatial} employ differentiable barriers to keep actions within certified safe sets. \cite{goodall2024leveraging} extend shielding methods to continuous domains by leveraging approximate dynamics models, enabling probabilistic safety guarantees during exploration.
\cite{selim2022safe} leveraged model-based RL and offline collected data to develop reachability-based safety layers to ensure safe actions for navigation scenarios. \cite{thananjeyan2021recovery, zhang2023safe} assumes access to an offline dataset for pretraining a cost critic along with a recovery policy, which is then fixed during online learning, limiting its applicability in settings where collecting sufficient offline data is challenging or costly. \textbf{While} our method also modifies actions to maintain safety, it differs fundamentally by relying on online-learned cost predictions rather than external models or handcrafted safe sets, and by smoothly scaling actions instead of hard blocking or overwriting them, preserving the agent’s exploratory behavior.

\textbf{Constrained RL Approaches.} Lagrangian-based algorithms~\cite{achiam2017constrained,ray2019benchmarking,stooke2020responsive} optimize dual formulation balancing rewards and costs, while budgeted RL~\cite{sootla2022saute,sootla2022enhancing} include the remaining cost budget in the state representation, allowing the agent to adapt its behavior based on how much cost it can afford. Risk-sensitive formulations, such as CVaR-CPO~\cite{zhang2024cvar}, constrain the conditional value-at-risk of cumulative costs, ensuring attention to costly violations. 
Reachability-based methods like RESPO~\cite{ganai2023iterative} estimate the probability of reaching safe regions and optimize policies to satisfy constraints or recover when outside the feasible set.
Bi-level optimization frameworks such as SRCPO~\cite{kim2024spectral} address the nonlinearity of risk measures by optimizing over dual variables, achieving strong constraint satisfaction in continuous control tasks.
Safety Editor~\cite{yu2022towards} trains two  Soft Actor-Critic (SAC)~\cite{haarnoja2018soft} agents: a utility maximizer and a safety editor that modifies unsafe actions, allowing it to fully overwrite the original action when necessary.
\textbf{Compared }to these methods, our approach offers a lightweight, modular alternative: instead of embedding constraints into the policy loss, relying on delicate dual updates, or training a second actor to overwrite unsafe actions, we regulate actions externally using learned cost critics. This continuous scaling preserves exploration while enabling seamless integration into standard off-policy RL pipelines.
\vspace{-13pt}
\begin{figure*}[t!]
    \centering
    \includegraphics[width=0.7\linewidth]{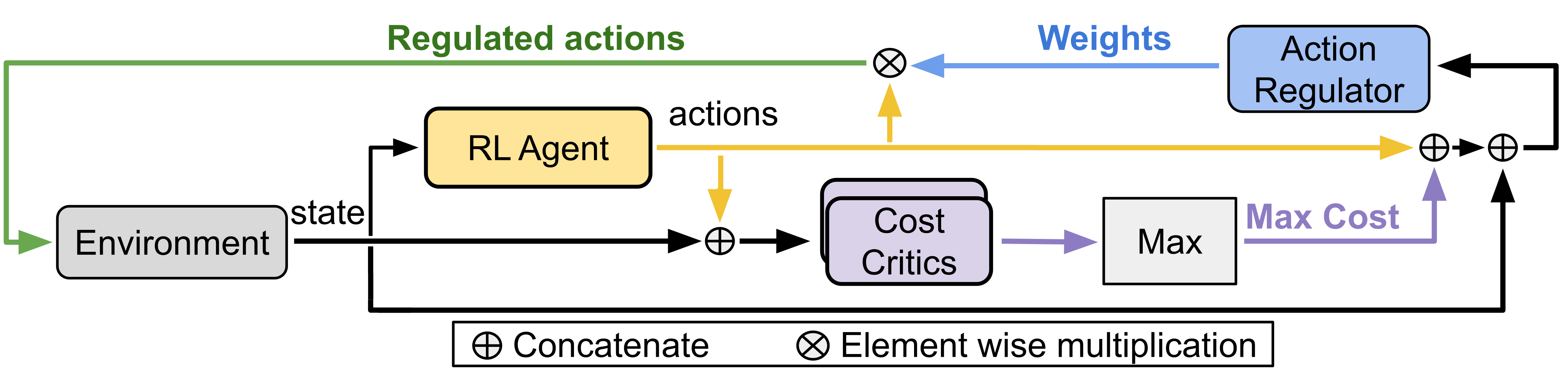}
    \caption{Overview of our modular safe RL architecture. The regulator~(blue) scales actions produced by the unconstrained RL agent~(yellow) based on predicted cost~(purple), producing safety-aware actions~(green) that are executed in the environment.}
    \label{fig:pipeline_simple}
    \vspace{-20pt}
\end{figure*}
\section{Preliminaries}
\label{sec:prelim}

\textbf{Markov Decision Processes.}
We consider a Markov Decision Process (MDP) defined by the tuple 
$\mathcal{M} = \langle \mathcal{S}, \mathcal{A}, P, r, \gamma \rangle$, 
where $\mathcal{S}$ is the state space, $\mathcal{A}$ the action space, 
$P(s'|s,a)$ the transition probability, 
$r: \mathcal{S} \times \mathcal{A} \rightarrow \mathbb{R}$ the reward function, 
and $\gamma \in (0,1)$ the discount factor. 
We assume continuous state and action spaces with 
$\mathcal{S} \subseteq \mathbb{R}^n$ and $\mathcal{A} \subseteq \mathbb{R}^d$.


\textbf{Constrained Markov Decision Processes.}
A CMDP~\cite{altman2021constrained} augments the MDP with a cost function 
$c: \mathcal{S} \times \mathcal{A} \rightarrow \mathbb{R}_{\geq 0}$ 
that quantifies safety violations. 
The objective is to maximize return while keeping the expected cumulative cost below a budget $\chi$:
\begin{equation}
\max_\pi\; \mathbb{E}_{s \sim d_0,\, a \sim \pi(\cdot \mid s)}[Q^\pi(s,a)]
\quad \text{s.t.}\; \mathbb{E}_{s \sim d_0,\, a \sim \pi(\cdot \mid s)}[Q_c^\pi(s,a)] \le \chi,
\label{eqn:CMDP} \tag{CMDP}
\end{equation}


\textbf{Cost Budget.}
The budget $\chi$ specifies the maximum allowable expected cumulative cost 
and is typically treated as a human-selected threshold that reflects task-specific safety requirements \cite{stooke2020responsive}. 
In this work, we assume a stricter setting by eliminating the cost budget, i.e., setting $\chi = 0$, 
similar to~\cite{ganai2023iterative}. This corresponds to a hard-safety regime that aims to achieve 
minimal constraint violations during learning.

\textbf{Problem Setting.}
With this stricter formulation, the problem considered in this work is to learn a policy that maximizes task rewards while minimizing constraint violations under the hard-safety regime $\chi = 0$. 
Formally, our objective reduces to:
\begin{equation}
\pi^* = \arg\max_\pi\; \mathbb{E}_{s \sim d_0,\, a \sim \pi(\cdot \mid s)}[Q^\pi(s,a)]
\quad \text{s.t.}\; \mathbb{E}_{s \sim d_0,\, a \sim \pi(\cdot \mid s)}[Q_c^\pi(s,a)] = 0.
\end{equation}
This assumption eliminates any positive cost budget and focuses on policies that aim to achieve minimal safety violations during training and execution.
\vspace{-10pt}

\section{Methodology}

We propose a modular safe reinforcement learning framework that regulates the actions of a task policy to reduce expected constraint violations without overriding agent decisions. The key idea is to scale actions based on their predicted cost, preserving exploration while inducing smoother and safer transitions in the environment.
\vspace{-10pt}

\subsection{Split Architecture for Reward and Cost Optimization}

Optimizing for both task rewards and safety constraints within a single policy often leads to instability or overly conservative behavior \cite{stooke2020responsive}. To address this, we decouple the reward and cost learning objectives across two modules: The \textbf{task policy} \( \pi_\phi(a|s) \), which is trained to maximize expected rewards without incorporating safety constraints.
The \textbf{regulator network} \( \rho_\theta(s, a, \hat{c}) \), which learns to scale the policy's actions based on cost predictions to minimize constraint violations.
\vspace{-10pt}

\subsection{Action Modulation via Regulator Scaling}
\label{sec:monotone}
In continuous control environments without stochasticity, the system evolves under deterministic transition dynamics of the form \( s_{t+1} = f(s_t, a_t) \), where \( s_t \in \mathcal{S} \) is the current state and \( a_t \in \mathcal{A} \subset \mathbb{R}^d \) is a \( d \)-dimensional real-valued action vector, with \( d \) denoting the number of action dimensions. Since actions directly control the system evolution, high-magnitude or poorly directed actions can result in constraint violations or unstable behaviors.
To mitigate this, we introduce a \emph{regulator network} \( \rho_\theta: \mathcal{S} \times \mathcal{A} \times \mathbb{R} \rightarrow (0,1]^d \), which learns a scaling vector with an individual factor for each action dimension based on the current state, the raw action, and its predicted cost. 
At each step, the agent samples a raw action \( a_t \sim \pi_\phi(\cdot | s_t) \), computes the cost estimate: \( \hat{c}_t = \max(Q_c^1(s_t, a_t), Q_c^2(s_t, a_t)) \) from a twin-critic architecture, and the regulator network outputs a scaling vector \( \rho_t \), see Fig. \ref{fig:pipeline_simple}. The final action applied to the system is:
\begin{equation}
    \tilde{a}_t = \rho_t \odot a_t, \quad \text{where } \rho_t = \rho_\theta(s_t, a_t, \hat{c}_t),
\end{equation}
where \( \odot \) denotes element-wise multiplication; each component of the action vector is multiplied by a scaling factor between 0 (applied when high risk is predicted, resulting in large attenuation) and 1 (applied when the action is predicted to be safe, resulting in no attenuation), smoothly reducing potentially unsafe actions proportional to predicted risk.
This element-wise modulation attenuates each component of the action based on its risk profile, reducing the magnitude of high-risk components. Unlike hard safety constraints that may override agent behavior, this approach preserves the agent's exploration behavior and allows stable off-policy learning.

\textbf{Assumptions.} 
Our scaling mechanism assumes: (i) \textit{Monotonicity}, where $Q_c(s, \rho \odot a)$ is expected to decrease as $\rho \to 0$, and (ii) \textit{Safety of Inaction}, where zero-magnitude actions are safer than high-magnitude ones. These properties hold in many robotic domains where safety costs often scale with action magnitude, such as high torques causing actuator wear or large contact forces risking joint damage. By designing costs with this structure, scaling becomes an intuitive and effective tool for enforcing safety.
\vspace{-8pt}

\subsection{Learning Objectives and Updates}
\textbf{Reward Learning.}  
We adopt a general off-policy reinforcement learning framework where the agent's actor and critic are trained using the scaled action \( \tilde{a}_t \), as this is the action that is actually executed in the environment. The reward critic is updated using:
\begin{align}
Q_r(s_t, \tilde{a}_t) \;\leftarrow\;&\; r(s_t, \tilde{a}_t)
+ \gamma \, \mathbb{E}_{\substack{s_{t+1} \sim p(\cdot | s_t, \tilde{a}_t)  a_{t+1} \sim \pi(\cdot | s_{t+1})}}
    \!\left[ Q_r(s_{t+1}, \tilde{a}_{t+1}) \right],
\label{eq:reward_bellman}
\end{align}

where \( \tilde{a}_{t+1} = \rho_{t+1} \odot a_{t+1} \) and \( \rho_{t+1} = \rho_\theta(s_{t+1}, a_{t+1}, Q_c(s_{t+1}, a_{t+1})) \).  
The policy is updated to maximize the expected return under the regulated action:
\begin{equation}
\mathcal{L}_{\text{actor}} = \mathbb{E}_{s_t \sim \mathcal{D}, a_t \sim \pi(\cdot | s_t)} \left[ - Q_r(s_t, \tilde{a}_t) \right],
\label{eq:actor_loss}
\end{equation}
ensuring that policy learning reflects the actual dynamics induced by the regulated action \(\tilde{a}_t\).
Our framework is algorithm-agnostic and can be integrated with any off-policy actor-critic method. For entropy-regularized algorithms such as SAC, the corresponding entropy term may be included in the actor objective.
In our experiments, we demonstrate compatibility with both SAC and Twin Delayed DDPG~(TD3)\cite{fujimoto2018addressing}.

\textbf{Cost Learning.}
The cost critic is also trained on the scaled actions using a TD-style Bellman backup~\cite{sutton1998reinforcement}:
\begin{align}
Q_c(s_t, \tilde{a}_t) \;\leftarrow\;&\; c(s_t, \tilde{a}_t) 
+ \gamma \, \mathbb{E}_{\substack{s_{t+1} \sim p(\cdot | s_t, \tilde{a}_t) a_{t+1} \sim \pi(\cdot | s_{t+1})}}
   \!\left[ Q_c(s_{t+1}, \tilde{a}_{t+1}) \right],
\label{eq:cost_bellman}
\end{align}
This ensures the critic reflects the safety implications of the actual executed action \(\tilde{a}_t\).

\textbf{Regulator Objective.}
The regulator is trained to minimize the predicted cost of the executed action~\( \tilde{a}_t \), while avoiding degenerate solutions that collapse actions toward zero. Its loss function is given by:
\begin{equation}
\begin{aligned}
\mathcal{L}_{\text{reg}} 
= \; \mathbb{E}_{s_t \sim \mathcal{D},\, a_t \sim \pi(\cdot | s_t)} 
\Big[ \, & \beta \cdot Q_c(s_t, \tilde{a}_t)
 - \lambda \cdot \log \rho_\theta(s_t, a_t, \hat{c}_t) \, \Big].
\end{aligned}
\label{eq:reg_loss}
\end{equation}
where \( \beta, \lambda > 0 \) are trade-off parameters.
The first term encourages the regulator to scale down actions that lead to high predicted costs. However, without the second term, a trivial solution where \( \rho_\theta(s, a, \hat{c}) \rightarrow 0 \) would minimize this objective by collapsing all actions—halting the agent’s behavior entirely. To counteract this, the second term acts as a \textit{barrier penalty} that diverges as any element of the scaling vector approaches zero. It encourages the regulator to retain as much of the original action magnitude as possible, unless high predicted cost necessitates suppression. 


\textbf{Optimality Trade-Off.}
The regulator's training objective can be interpreted as solving a local constrained optimization problem at each state-action pair \( (s_t, a_t) \):
\begin{equation}
\min_{\rho \in (0,1]^d} \; \beta \cdot Q_c(s_t, \rho \odot a_t) - \lambda \cdot \log (\rho +\epsilon),
\label{eq:reg_opt_problem}
\end{equation}
where the logarithm is applied element-wise to the scaling vector \( \rho \), and we include \( \epsilon \) to avoid instability as \( \rho \to 0 \), ensuring gradients remain well-defined during training.

The coefficients \( \beta \) and \( \lambda \) balance the trade-off between minimizing predicted cost and preserving action magnitude: larger \( \lambda \) encourages less suppression, while larger \( \beta \) prioritizes cost reduction. 
Since \( Q_c(s, \rho \odot a_t) \) is typically a nonlinear function of the scaled action, the optimization problem lacks a closed-form solution but can be efficiently solved via gradient-based updates.
This formulation ensures that the regulator selectively attenuates risky action dimensions while retaining as much of the agent’s original behavior as possible.

\textbf{Gradient Flow and Modularity.}  
To ensure clean modularity, we detach the scaling weights \( \rho_\theta(s_t, a_t, \hat{c}_t) \) from the computational graph when updating both the reward and cost critics, preventing gradients from flowing through the regulator. Similarly, the actor receives no gradients from the regulator, learning purely from task returns. Moreover, the regulator is updated independently via its own objective, ensuring that reward maximization and safety modulation remain decoupled.

This design is particularly well-suited for \textbf{off-policy reinforcement learning}, where updates are performed using transitions stored in a replay buffer, independent of the current policy. Since the regulator modulates actions \emph{after} sampling from the policy \( \pi_\phi(\cdot \mid s) \), the executed action \( \tilde{a} = \rho_\theta(s, a, \hat{c}) \odot a \) differs from the originally sampled action \( a \), and only the regulated action is stored and used for training. Off-policy methods naturally accommodate this, as policy and critic updates rely on the actual executed actions rather than the distribution used to generate them.

\textbf{Implementation Details.}
The RL agent follows its baseline implementation without modification.
The regulator and twin cost critics are feedforward neural networks with two hidden layers of 256 units and ReLU activations. 
The regulator outputs element-wise scaling factors $\rho \in (0,1]^d$  via a sigmoid activation and is trained using the twin cost critics' predictions. The code is provided online\footnote{https://github.com/HumanoidsBonn/Constraint-Aware-Adaptive-Action-Scaling}.
\vspace{-7pt}

\section{Experiments}
\label{sec:results}
\begin{figure}[t]
    \centering
    \includegraphics[width=0.9\linewidth]{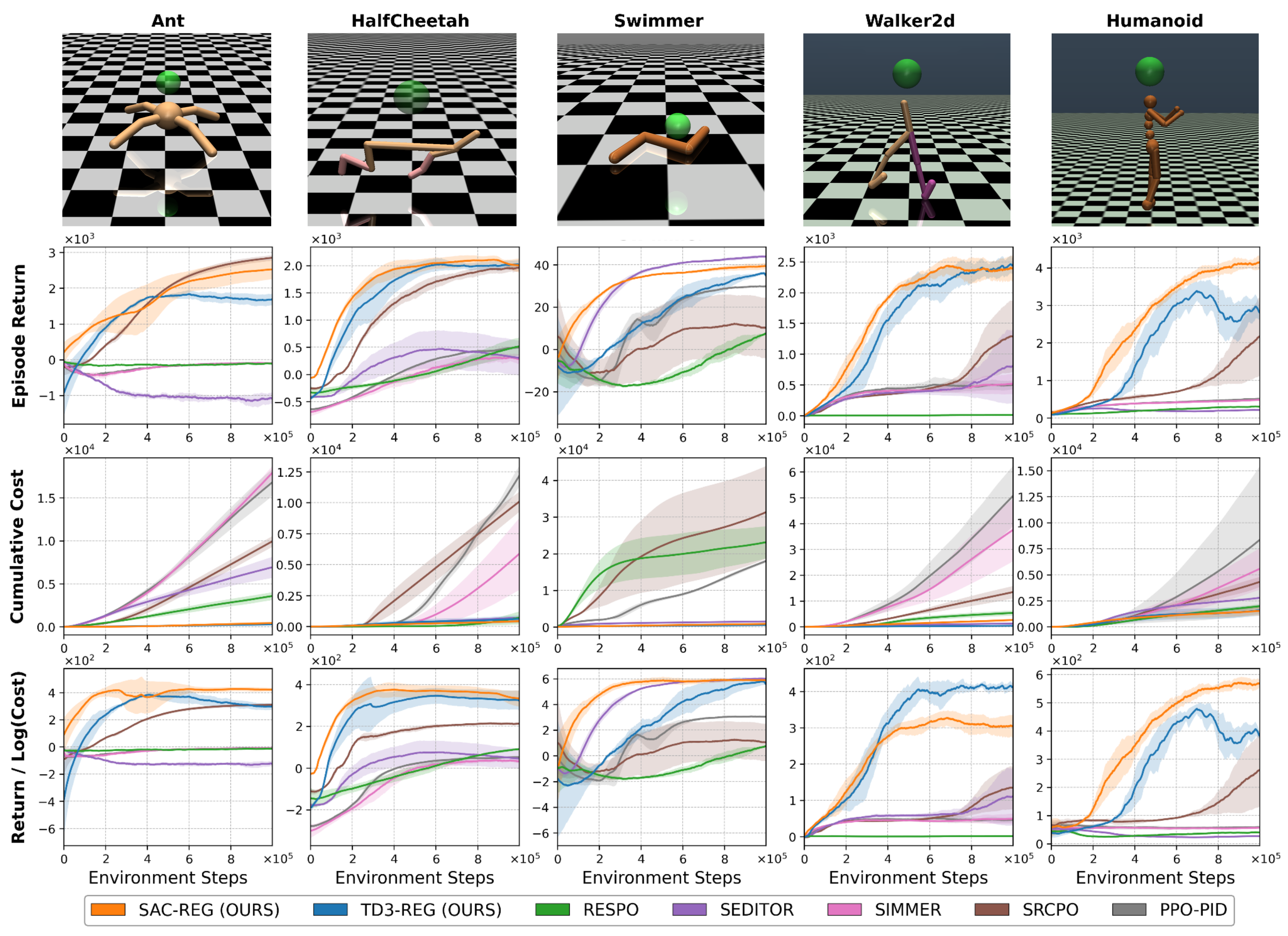}
    \caption{
    Performance comparison on the Safety Gymnasium locomotion environments.
    Each method is averaged over three independent runs; bold lines indicate the mean, and shaded areas show the standard deviation.
    Our methods ({SAC-REG} and {TD3-REG}) consistently achieve the best trade-off between return and cumulative constraint cost across all environments. 
    \textbf{Top:} Episode return. \textbf{Middle:} Cumulative safety cost. \textbf{Bottom:} Return-to-Log-Cost ratio. 
    Our methods outperform strong baselines, including {PPO-PID}~\cite{stooke2020responsive}, {SIMMER}~\cite{sootla2022enhancing}, {SRCPO}~\cite{kim2024spectral}, {RESPO}~\cite{ganai2023iterative}, and {SEDITOR}~\cite{yu2022towards}. SIMMER is omitted from the \texttt{Swimmer} plot as it consistently yields negative returns, moving opposite to the target velocity.}
    \label{fig:results_comparison_baselines}
    \vspace{-20pt}
\end{figure}

Our experiments are designed to achieve the following objectives:
(i) compare our approach against state-of-the-art safe RL baselines across different dynamical systems,
(ii) analyze the influence of the key hyperparameters (\(\lambda\) and~\(\beta\)) from Eq.~\ref{eq:reg_opt_problem}, which govern the trade-off between action preservation and cost suppression, 
(iii) evaluate the regulator’s action-scaling mechanism through ablations such as element-wise versus scalar regulation on different systems, and 
(iv) study robustness under injected sensor and actuator noise, highlighting the method’s potential for sim-to-real transfer.

\textbf{Environments:}  
We evaluate our method on locomotion tasks from the \textbf{Safety Gym} benchmark~\cite{ji2023safety}, namely \texttt{Ant}, \texttt{Walker2d}, \texttt{Swimmer}, \texttt{HalfCheetah}, and \texttt{Humanoid}. 
In these velocity tasks, a safety cost is incurred whenever the center-of-mass speed exceeds a predefined threshold. 
Because the cost signal is sparse and triggered only by such threshold violations, these environments provide a challenging setting for safe RL, while naturally aligning with our regulator’s goal of attenuating large actions that are most likely to induce violations. 

\textbf{Baselines:}  
We compare our regulator against five state-of-the-art Safe RL baselines. \textbf{PPO-PID}~\cite{stooke2020responsive} augments PPO with a PID-controlled Lagrangian multiplier to mitigate instabilities commonly observed in dual updates during constrained optimization. \textbf{Simmer}~\cite{sootla2022enhancing} augments PPO with a safety state that tracks the remaining safety budget. \textbf{Safety Editor}~\cite{yu2022towards} uses two SAC agents, one for maximizing the reward and another for editing unsafe actions. \textbf{RESPO}~\cite{ganai2023iterative} estimates reachability sets and constrains policy to remain within safe regions. \textbf{SRCPO}~\cite{kim2024spectral} formulates a bi-level constrained optimization using spectral risk measures to achieve a near-zero constraint violation rate while maximizing reward. 

\textbf{Metrics: }
Similar to prior safe RL studies, we report returns and cumulative costs as in~\cite{thananjeyan2021recovery, ganai2023iterative, goodall2024leveraging}, and follow~\cite{thananjeyan2021recovery} in using the return-to-cost (RC) ratio to capture the trade-off between task performance and safety. We measure (i) episodic return, (ii) cumulative cost during training, which reflects the total number of constraint violations and, in sparse-cost settings such as Safety Gym velocity tasks, implicitly captures violation frequency, and (iii) the RC ratio, defined as the total return divided by cumulative cost. For visualization, we plot the return divided by the logarithm of the accumulative cost, which improves interpretability of the safety-performance trade-off.
\vspace{-13pt}

\subsection{Comparison Against Baselines:}
\paragraph{Safety Gym Results.}
Across the locomotion tasks in the Safety Gymnasium suite, our methods—\textbf{SAC-Regulator} and \textbf{TD3-Regulator}—consistently deliver strong task performance while substantially reducing safety violations. Each method was evaluated over three random seeds; bold lines in Fig.\ref{fig:results_comparison_baselines} denote the mean return across runs, with shaded regions representing standard deviation. Compared to the baselines, our approach achieves higher or comparable episode returns, indicating that soft action scaling does not hinder exploration. Notably, \textbf{TD3-Regulator} achieves the greatest cost reductions, with up to \textbf{126×} lower cumulative cost in \texttt{Walker2d}, \textbf{64×} in \texttt{Ant}, and \textbf{86×} in \texttt{Swimmer}. Meanwhile, \textbf{SAC-Regulator} outperforms both \textbf{TD3-Regulator} and all baselines in \texttt{HalfCheetah} and \texttt{Humanoid}, achieving cost reductions of up to \textbf{28×} and \textbf{5×}, respectively.
RESPO can reach comparable returns when trained for 9M steps, but only with substantially higher violations and failure to converge in \texttt{Humanoid}.

\begin{table}[t]
\centering
\scriptsize
\resizebox{0.7\columnwidth}{!}{%
\begin{tabular}{lccccc}
\toprule
\multicolumn{6}{c}{\textbf{Relative Return Improvement of SAC-Reg Over Baselines (↑)}} \\
\midrule
\textbf{Method} & Ant & HalfCheetah & Swimmer & Walker2d & Humanoid \\
\midrule
PPO-PID~\cite{stooke2020responsive} & 27.33 & 3.04   & 0.32    & 3.76   & 7.13  \\
SIMMER~\cite{sootla2022enhancing}   & 26.14 & 6.09   & 1.21    & 3.59   & 7.60  \\
SEditor~\cite{yu2022towards}        & 3.34  & 5.64   & -0.10   & 1.95   & 18.52 \\
RESPO~\cite{ganai2023iterative}     & 23.97 & 2.89   & 0.17    & 204.40 & 12.73 \\
SRCPO~\cite{kim2024spectral}        & -0.11 & 0.02   & 2.90    & 0.86   & 0.90  \\
\midrule
\multicolumn{6}{c}{\textbf{Relative Cost Compared to SAC-Reg (↓)}} \\
\midrule
PPO-PID~\cite{stooke2020responsive} & 39.29 & 28.39 & 21.31 & 19.11 & 5.46 \\
SIMMER~\cite{sootla2022enhancing}   & 41.88 & 13.70 & 54.22 & 14.14 & 3.64 \\
SEditor~\cite{yu2022towards}        & 16.19 & 1.67  & 1.77  & 0.48  & 1.88 \\
RESPO~\cite{ganai2023iterative}     & 8.36  & 1.65  & 27.39 & 2.03  & 1.29 \\
SRCPO~\cite{kim2024spectral}        & 23.24 & 23.55 & 37.10 & 5.07  & 2.82 \\
\bottomrule
\end{tabular}%
}
\caption{Relative return improvement (↑) and relative cumulative cost (↓) of SAC-Reg compared to baselines across locomotion tasks. SAC-Reg consistently achieves higher returns and lower cumulative costs than prior safe RL methods across all environments.}
\label{tab:return_cost_sacreg}
\vspace{-12pt}
\end{table}

\begin{table}[t]
\centering
\scriptsize
\resizebox{0.7\columnwidth}{!}{%
\begin{tabular}{lccccc}
\toprule
\multicolumn{6}{c}{\textbf{Relative Return Improvement of TD3-Reg Over Baselines (↑)}} \\
\midrule
\textbf{Method} & SafetyAnt & HalfCheetah & Swimmer & Walker2d & Humanoid \\
\midrule
PPO-PID~\cite{stooke2020responsive} & 18.49 & 3.05  & 0.17   & 3.87   & 4.51 \\
SIMMER~\cite{sootla2022enhancing}   & 17.70 & 6.11  & 1.18   & 3.70   & 4.83 \\
SEditor~\cite{yu2022towards}        & 2.55  & 5.65  & -0.20  & 2.02   & 12.22 \\
RESPO~\cite{ganai2023iterative}     & 16.26 & 2.90  & 3.59   & 209.23 & 8.30 \\
SRCPO~\cite{kim2024spectral}        & -0.41 & 0.02  & 2.46   & 0.91   & 0.29 \\
\midrule
\multicolumn{6}{c}{\textbf{Relative Cost Compared to TD3-Reg (↓)}} \\
\midrule
PPO-PID~\cite{stooke2020responsive} & 60.14 & 19.77 & 34.06 & 126.18 & 5.20 \\
SIMMER~\cite{sootla2022enhancing}   & 64.11 & 9.54  & 86.65 & 93.37  & 3.47 \\
SEditor~\cite{yu2022towards}        & 24.78 & 1.16  & 2.84  & 3.15   & 1.74 \\
RESPO~\cite{ganai2023iterative}     & 12.80 & 1.15  & 43.78 & 13.42  & 1.23 \\
SRCPO~\cite{kim2024spectral}        & 35.57 & 16.39 & 59.28 & 33.50  & 2.69 \\
\bottomrule
\end{tabular}%
}
\caption{Relative return improvement (↑) and relative cumulative cost (↓) of TD3-Reg compared to baselines across locomotion tasks.TD3-Reg demonstrates similar trends, outperforming baselines in return while maintaining substantially lower cumulative costs.}
\vspace{-15pt}
\label{tab:return_cost_td3reg}
\end{table}

Tables~\ref{tab:return_cost_sacreg} and~\ref{tab:return_cost_td3reg} summarize results for both regulators against established baselines. Two metrics are reported: the \textit{relative return improvement},  
\[
\frac{\text{Return}_{\text{Ours}} - \text{Return}_{\text{Baseline}}}{|\text{Return}_{\text{Baseline}}|},
\]  
and the \textit{relative cumulative cost} of each baseline normalized by our method. Positive return values indicate improved task performance, while cumulative cost ratios above 1.0 indicate higher constraint violations than ours. For example, in \texttt{Walker2d}, \textbf{SAC-Reg }outperforms RESPO with a relative return improvement of 204.4, corresponding to a 20,440\% increase. 

Overall, our methods deliver the lowest cumulative cost across all tasks without compromising return. Unlike approaches such as \textsc{SEditor} or \textsc{RESPO}, which improve safety at the expense of performance, our regulators preserve exploration and consistently achieve superior return-to-cost trade-offs. This demonstrates the effectiveness and generality of decoupling safety regulation from reward learning.


\begin{figure*}[t]
    \centering
    \includegraphics[width=0.75\linewidth]{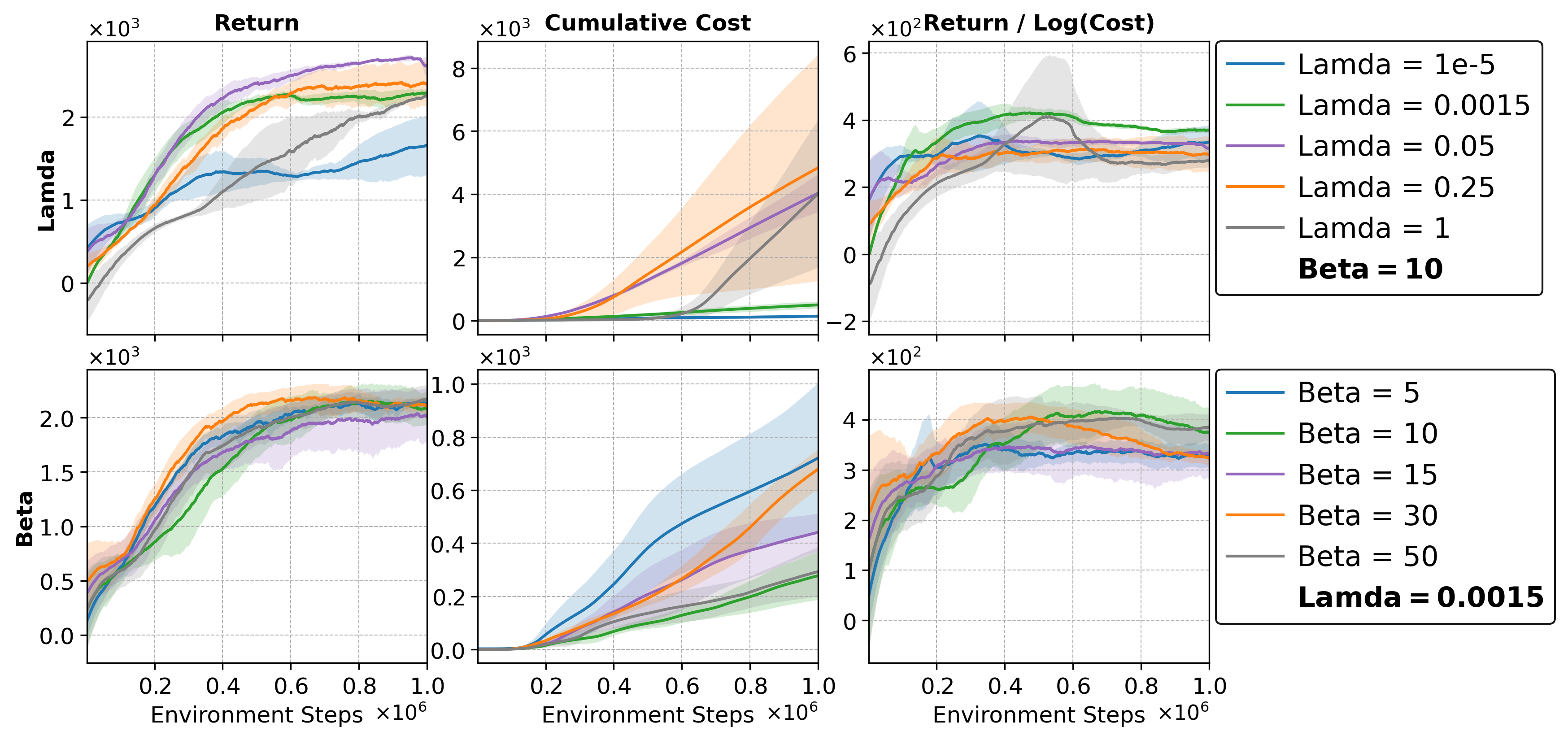}
    \caption{Ablation Study for evaluating the impact of the regulator hyperparameters 
    \(\lambda\) and \(\beta\) on Return, Cumulative Cost, and Return-to-Cost ratio. 
    Each curve shows the mean across three runs, and shaded regions indicate standard 
    deviation. The top row varies \(\lambda\) with fixed \(\beta = 10\); the bottom row 
    varies \(\beta\) with fixed \(\lambda = 0.0015\). As seen, smaller \(\lambda\) values reduce cumulative cost, with 
\(\lambda = 0.0015\) giving the best balance between performance and safety, while \(\beta = 10\) provides the most favorable trade-off overall.}
    \label{fig:ablation_results}
    \vspace{-13pt}
\end{figure*}

\textbf{Ablation Study on \(\lambda\) and \(\beta\).} 
We conduct an ablation study in the \texttt{Ant} environment to evaluate the sensitivity of our regulator framework to the hyperparameters \(\lambda\) and \(\beta\), which control the trade-off between action retention and cost suppression. When varying \(\lambda\) over the range \(\{1\times 10^{-5},\ 0.0015,\ 0.05,\ 0.25,\ 1.0\}\), we find that smaller values lead to significantly lower cumulative costs. In particular, \(\lambda = 0.0015\) achieves the best balance between constraint satisfaction and task performance. Higher values of \(\lambda\) result in larger action magnitudes and consequently higher constraint violations. Similarly, varying \(\beta\) over \(\{5,\ 10,\ 15, \ 30,\ 50\}\) shows that \(\beta = 10\) achieves the best overall safety-performance balance, minimizing constraint violations while maintaining high return. These results, as shown in Fig.~\ref{fig:ablation_results}, highlight the importance of properly tuning the regulator’s loss coefficients to achieve optimal return-to-cost behavior.

\begin{figure*}[t]
    \centering
    \includegraphics[width=0.8\linewidth]{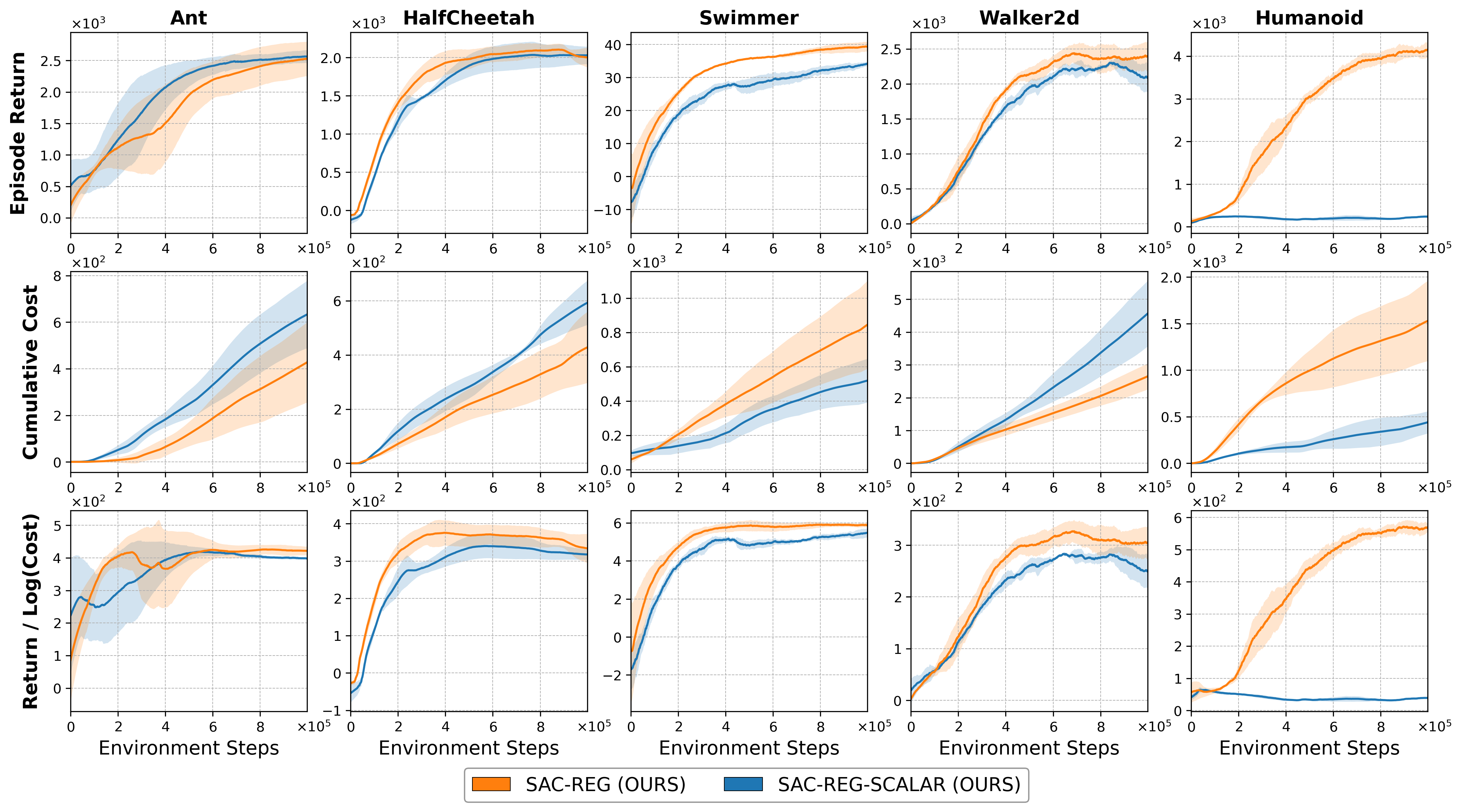}
        \caption{Comparison between element-wise and scalar action scaling. While both perform similarly in most tasks, the scalar variant fails to converge in the high-dimensional \texttt{Humanoid} environment, indicating that element-wise scaling improves safety and stability in complex control settings.}
        \vspace{-5pt}
    \label{fig:act_scales}
\end{figure*}

\textbf{Element-wise vs. Scalar Regulation:}
To evaluate the impact of element-wise action regulation, we conducted an ablation study comparing our full regulator with a simplified variant that uses a single scalar value to uniformly scale all action dimensions. Figure~\ref{fig:act_scales} presents results across all Safety Gymnasium locomotion tasks. While the scalar variant achieves comparable performance in most environments, it fails to converge in the high-dimensional \texttt{Humanoid} task. This suggests that element-wise scaling is particularly important in complex control settings, where individual action dimensions exhibit distinct risk profiles. Fine-grained modulation allows the regulator to target risky joints more precisely, improving both safety and learning stability.

\begin{table}[t]
\centering
\scriptsize
\resizebox{0.6\columnwidth}{!}{%
\begin{tabular}{c|cc|cc|cc|cc}
\toprule
\multirow{2}{*}{\textbf{Step}} 
& \multicolumn{2}{c|}{\textbf{Noise 0.00}} 
& \multicolumn{2}{c|}{\textbf{Noise 0.025}} 
& \multicolumn{2}{c|}{\textbf{Noise 0.05}} 
& \multicolumn{2}{c}{\textbf{Noise 0.10}} \\
\cmidrule(lr){2-3} \cmidrule(lr){4-5} \cmidrule(lr){6-7} \cmidrule(lr){8-9}
& Return & Cost & Return & Cost & Return & Cost & Return & Cost \\
\midrule
100k  & 1961 & 6   & 753  & 0   & 762  & 28  & 677  & 1   \\
300k  & 2558 & 79  & 1554 & 78  & 1614 & 165 & 1103 & 113 \\
500k  & 2492 & 199 & 1945 & 195 & 1969 & 282 & 1381 & 387 \\
700k  & 2533 & 282 & 2139 & 230 & 2071 & 339 & 1596 & 572 \\
900k  & 2501 & 388 & 2104 & 262 & 2074 & 479 & 1589 & 767 \\
1000k & 2571 & 412 & 2016 & 312 & 2089 & 533 & 1510 & 849 \\
\bottomrule
\end{tabular}}
\caption{Training performance under different levels of injected Gaussian noise 
($\sigma = 0.00, 0.025, 0.05, 0.10$) in observations and actions. 
Values show episode return and cumulative cost at checkpoints. 
Our regulator maintains bounded costs across noise levels, demonstrating robustness relevant for sim-to-real transfer.}
\vspace{-12pt}
\label{tab:noise}
\end{table}

\textbf{Robustness and Sim-to-Real Transfer.}
To approximate uncertainties encountered on physical robots, we inject Gaussian noise into both observations and actions during training, modeling sensor measurement errors and actuator execution noise. 
Agents are trained with noise levels ($\sigma = 0, 0.025, 0.05, 0.10$), and the resulting training performance is summarized in Table~\ref{tab:noise}. 
Across all noise settings, our regulator achieves strong returns while keeping cumulative costs bounded. 
Even under the highest noise level ($\sigma = 0.10$), performance remains stable, highlighting robustness to sensing and actuation imperfections and supporting the method’s potential for sim-to-real transfer.

\section{Conclusion}

We introduced a modular and practical framework for safe reinforcement learning that decouples reward maximization from safety enforcement through a cost-aware regulator. Instead of overriding agent actions, our method scales them smoothly based on predicted constraint violations, preserving exploration and enabling stable off-policy learning. The regulator uses twin cost critics for robust cost estimation and is trained with a loss that balances risk reduction and action preservation. Our approach is model-free and integrates seamlessly with existing off-policy RL pipelines. Empirical results on diverse benchmarks demonstrate that our method consistently achieves the highest return-to-cost ratios, reducing constraint violations by up to 126 times while maintaining or improving task performance relative to prior state-of-the-art methods.
The regulator aligns with real-world safety limits such as torque bounds in manipulators, and joint load management in legged robots. Robustness experiments with injected observation and action noise further demonstrate bounded costs and stable returns under uncertainty, supporting the potential for sim-to-real transfer. A key direction for future work is to develop principled strategies for automatically tuning the regulator hyperparameters ($\lambda$ and $\beta$) and to extend the approach beyond input-magnitude costs toward more general safety constraints.
\section*{Acknowledgments}
M. Dawood, U. Ahmed Siddiquie, S. Khorshidi, and M. Bennwitz are with the Humanoid Robots Lab and the Center for Robotics, University of Bonn, Germany. Additionally, M. Dawood and M. Bennwitz are with the Lamarr Institute for Machine Learning and Artificial Intelligence. This work has partially been funded by the German Federal Ministry of Research, Technology and Space (BMFTR) under the Robotics Institute Germany (RIG).

\bibliography{bibliography}

\end{document}